\documentclass[letterpaper, 10 pt, conference]{ieeeconf}  
\usepackage{graphicx}
\usepackage{float} 
\usepackage{subfigure}

\IEEEoverridecommandlockouts                              

\usepackage{graphics} 
\usepackage{epsfig} 
\usepackage{mathptmx} 
\usepackage{times} 
\usepackage{amsmath} 
\usepackage{amssymb}  
\usepackage{multirow}
\usepackage{cite}    
\makeatletter
\let\NAT@parse\undefined
\makeatother
\usepackage{hyperref}
\hypersetup{
colorlinks=true,
linkcolor=black,
citecolor=black
}

\usepackage{array}
\usepackage{booktabs}
\usepackage{caption}
\usepackage{ragged2e}
\usepackage{diagbox}
\usepackage{makecell}
\usepackage{booktabs} 
\usepackage{multirow} 
\usepackage{url}

\title{\LARGE \bf
LangGrasp: Leveraging Fine-Tuned LLMs for Language Interactive Robot Grasping with Ambiguous Instructions
}

\author{Yunhan Lin, Wenqi Wu, Zhijie Zhang and Huasong Min$^{*}$
\thanks{This work is supported by the National Key R\&D Program of China (grant No.: 2022YFB4700400), National Natural Science Foundation of China (grant No.: 62073249), Key R\&D Program of Hubei Province (grant No.: 2023BBB011), and China Scholarship Council.}
\thanks{Yunhan Lin, Wenqi Wu, Zhijie Zhang and Huasong Min are with the School of Computer Science and Technology, Hubei Province Key Laboratory of Intelligent Information Processing and Real-time Industrial System, Institute of Robotics and Intelligent Systems, Wuhan University of Science and Technology, Wuhan, China.}%
\thanks{*Coresponding author: Huasong Min, email: mhuasong@wust.edu.cn.}%
}

\begin{document}

\maketitle
\thispagestyle{empty}
\pagestyle{empty}

\begin{abstract}

 The existing language-driven grasping methods struggle to fully handle ambiguous instructions containing implicit intents. To tackle this challenge, we propose LangGrasp, a novel language-interactive robotic grasping framework. The framework integrates fine-tuned large language models (LLMs) to leverage their robust commonsense understanding and environmental perception capabilities, thereby deducing implicit intents from linguistic instructions and clarifying task requirements along with target manipulation objects. Furthermore, our designed  point cloud localization module, guided by 2D part segmentation, enables partial point cloud localization in scenes, thereby extending grasping operations from coarse-grained object-level to fine-grained part-level manipulation. Experimental results show that the LangGrasp framework accurately resolves implicit intents in ambiguous instructions, identifying critical operations and target information that are unstated yet essential for task completion. Additionally, it dynamically selects optimal grasping poses by integrating environmental information. This enables high-precision grasping from object-level to part-level manipulation, significantly enhancing the adaptability and task execution efficiency of robots in unstructured environments. More information and code are available here: https://github.com/wu467/LangGrasp.

\end{abstract}

\section{INTRODUCTION}

With the increasing deployment of robots in daily environments, natural language based human-robot interaction and manipulation have significantly enhanced efficiency and task executability due to their intuitiveness and flexibility. However, accurately interpreting ambiguous or context dependent linguistic instructions in dynamic, unstructured environments, particularly those containing implicit intents, remains a critical challenge. Furthermore, existing language-guided grasping frameworks are limited to predicting object-level grasping poses, neglecting the functional distinctions between object parts. For instance, in tasks such as cutting fruit with a knife, successful execution requires not only recognition of the object's global geometry but also a functional-structural analysis of its parts to identify the handle as the optimal grasping region.
\begin{figure}[htbp]
    \centering
    \includegraphics[scale=0.052]{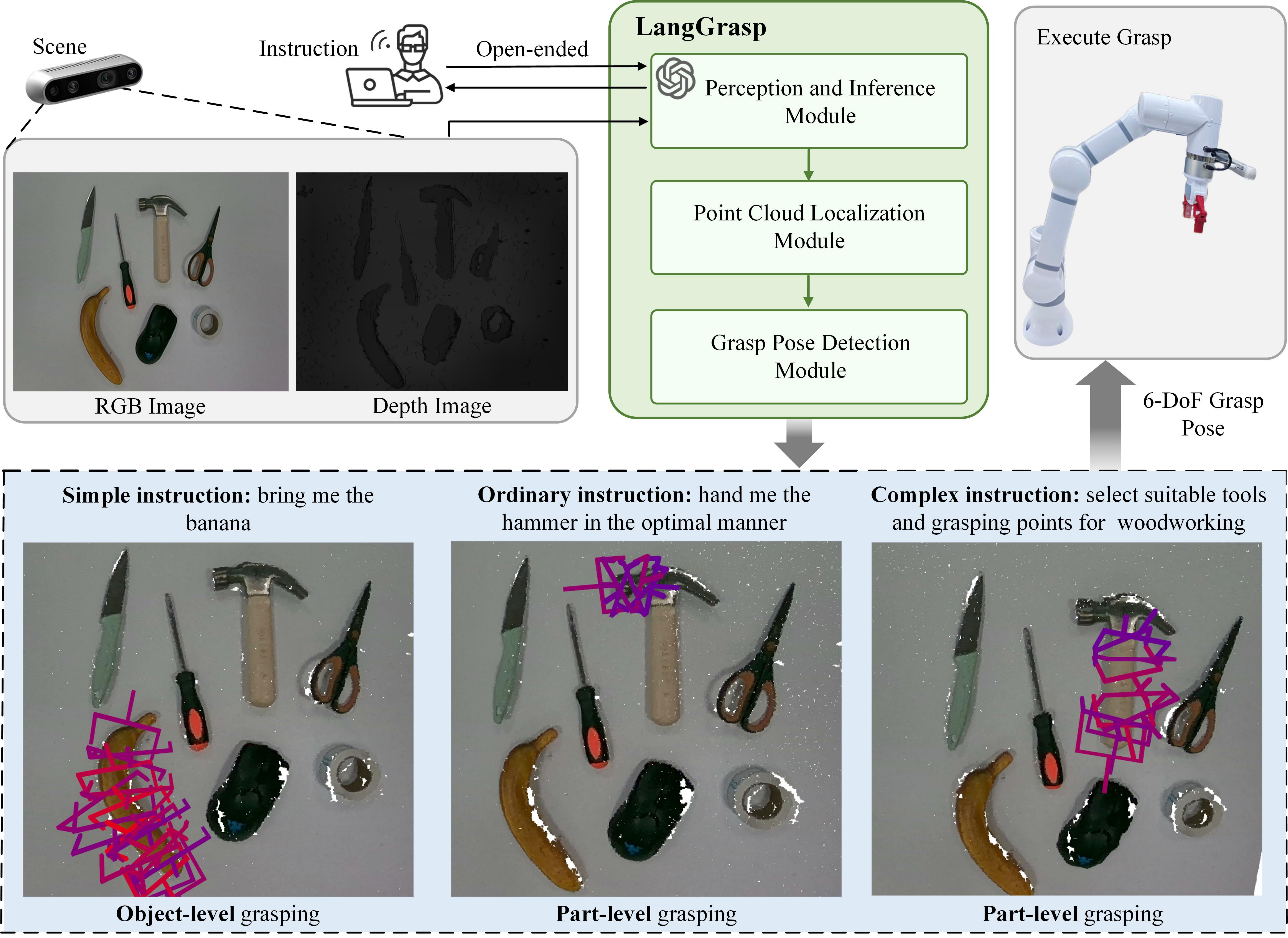}
    \caption{The overall framework of our proposed method. The LangGrasp framework primarily consists of three modules: Perception and Inference, Point Cloud Localization, and Grasp Pose Detection. Inputting an RGB image, a Depth image, and language instructions, the framework outputs optimal 6-DoF grasp poses.}
    \label{fig1}
    \vspace{-9pt} 
\end{figure}
This necessitates that robotic systems possess both efficient natural language understanding capabilities for parsing complex linguistic constructs and fundamental visual perception for fusing semantic-geometric cues, enabling commonsense-driven intent disambiguation. LLMs, leveraging their vast world knowledge, demonstrate significant potential in resolving linguistic ambiguity, contextual reasoning, and functional analysis. Integrating LLMs into human-robot interaction frameworks to enhance the comprehension of complex instructions and task planning efficiency has emerged as a prominent research topic.

Previous studies \cite{c1},\cite{c2} treat natural language as high-level semantic cues, combining visual information with pretrained visual grounding models to localize target object and determine grasping region. Such approaches rely on explicit semantic cues and unambiguous instruction structures, which results in their failure to handle linguistically ambiguous or polysemous expressions. Recent work \cite{c3} has integrated LLMs into this paradigm, utilizing their advanced natural language understanding and reasoning capabilities to tackle challenges in extracting target objects and task parameters from vague or inference-dependent instructions. However, due to the inherent uncertainty and redundancy in LLM outputs, even with optimized prompt engineering, are mains essential to extract actionable information, thus increasing operational complexity within the pipeline.

To address these challenges, we propose LangGrasp, a novel language interactive grasping framework, as illustrated in Figure \ref{fig1}. The framework leverages fine-tuning to preserve the LLM's vast world knowledge while significantly enhancing its reasoning and decision-making capabilities in open-ended interaction scenarios. During the interaction phase, the perception and inference module generates operation sequences in JSON format based on the environmental context and multi-turn dialogue history. Unlike unstructured text responses, structured output enables direct extraction of target objects and task specifications without additional parsing steps. At the operational level, we have designed a point cloud localization module to meet the precision requirements in open environments. By leveraging a pretrained 2D part segmentation model, our point cloud localization module achieves localization for target objects or their parts. When manipulating a composite object with multiple parts, the framework dynamically selects the optimal manipulation region by synthesizing geometric shapes, part attributes, and task requirements. Finally, the grasp pose detection module predicts the grasp pose for the localized point cloud.

As shown in Figure \ref{fig1}, LLM is used as the core of human-robot interaction in LangGrasp, with three instructions demonstrated. The first is a simple instruction: the robot is asked to bring a banana to the user, and it can select any grasping pose to pick up the banana. The second and third instructions generate part-level grasp poses, supporting long-horizon task execution by combining linguistic instructions with commonsense reasoning. Specifically, for the instruction ``Hand me the hammer'' the robot grasps the hammer’s head while orienting the handle toward the user. This commonsense-based strategy ensures the stability of the grasp and optimizes ergonomic comfort during the handover process. In the third case, when executing carpentry instructions that require hammer strikes, the robot selects the handle of the hammer to ensure the correct striking posture. These diverse strategies enhance adaptability to complex instructions, while also improving operational reliability and precision.

The main contributions of this paper are as follows:

(1) We propose LangGrasp, a novel language interactive grasping framework that leverages the commonsense reasoning and environmental perception capabilities of LLMs to interpret natural language instructions of varying complexity, enabling fine-grained analysis of target objects. Through the designed perception and inference module, point cloud localization module, and grasp pose detection module, LangGrasp achieves precise grasping region selection, advancing the grasping targets from the object-level to the part-level.

(2) We construct a fine-tuning dataset for LLMs specifically for robotic grasping tasks, enhancing the parsability and interpretability of LLM outputs when processing different complexities of linguistic instructions.

(3) Two platforms, a desktop experimental scene and a cabinet experimental scene, are designed to validate the efficiency of the proposed method. The experimental results demonstrate the advantages of proposed modules and the effectiveness of the LangGrasp framework.

\begin{figure*}[htbp]
    \centering
    \includegraphics[scale=0.057]{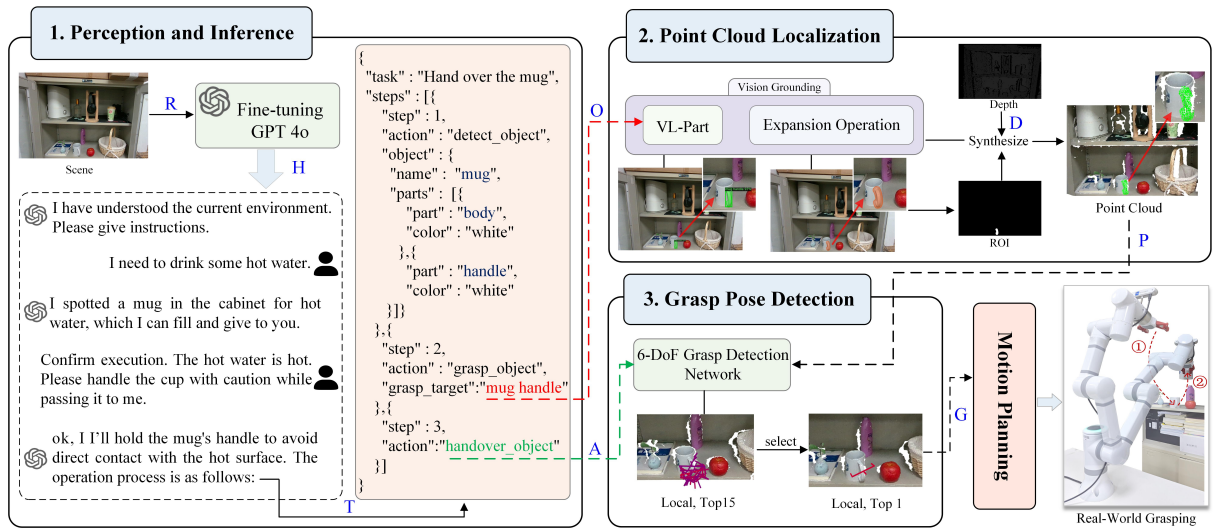}
    \caption{The procedure of LangGrasp consists of three stages: 1. the Perception and Inference stage, where structured reasoning results are generated based on the current scene and multi-turn dialogue information; 2. the Point Cloud Localization stage, where target object point clouds are localized in the global point cloud using semantic information generated in the previous stage, and the target point cloud region is optimized through an expansion strategy; 3. the Grasp Pose Detection stage, where the 6-DoF grasp pose of the local point cloud is predicted, and the pose with the highest score data is sent to the robotic arm for grasp execution.}
    \label{fig2}
    \vspace{-5pt} 
\end{figure*}

\section{RELATED WORK}
\subsection{LLMs for Robotic} 

Recent advancements in LLMs have revitalized robotics research. Several approaches \cite{c4},\cite{c5},\cite{c6} employ LLMs as task planners that decompose complex instructions into executable atomic action sequences, thereby bridging high-dimensional task spaces with low-dimensional control spaces. Our previous work \cite{c7} developed an LLM-behavior tree integration framework that dynamically updates behavior trees based on real-time environmental perception, enabling adaptive decision-making in complex, dynamic environments. The Code-as-Policies framework \cite{c8} introduced the concept of language model generated programs, leverages the code generation capabilities of LLMs to translate natural language commands into policy code. This approach utilizes predefined action function libraries and generalizes to novel task scenarios. However, due to the limited perception of physical environments, these methods are constrained to offering coarse-grained task planning solutions. With the emergence of multimodal large models, Huang et al. proposed Voxposer \cite{c9}, which integrates multi-view workspace images and linguistic instructions as input to a vision-language model. This framework constructs a 3D value map to guide the synthesis of robotic end-effector motion trajectories, enabling zero-shot generalization. Wang et al. introduced RT-2 \cite{c10}, which fine-tunes a vision-language model by incorporating discretized robotic actions, thereby transforming it into a vision-language-action model. This architecture enables the robot to directly output robotic actions based on visual perception and textual instructions, thereby advancing the implementation of perception-action loops. In this paper, we focus on leveraging the perception, reasoning, and planning capabilities of LLMs as the cognitive core for decision-making and interaction, establishing an open interaction framework that links users, LLMs, and robots. The proposed framework addresses the challenge of converting language instructions with varying complexities into task plans with clear intentions and executability, thereby enhancing the intelligence and robustness of human-robot collaboration.

\subsection{Language-guided Robotic Grasping}

Language driven grasping is a fundamental task in human-robot interaction, typically comprising two stages: (1) target object localization based on linguistic commands, and (2) grasp pose prediction for the identified object. Target object localization is a fundamental prerequisite for this task. Sun et al. \cite{c2} proposed CRIS, a CLIP-based segmentation model that integrates Mask R-CNN to jointly predict the spatial positions of target objects. Mohit et al. \cite{c11} introduced CLIPort, which integrates the spatial precision of Transporter networks and CLIP's semantic comprehension within a unified convolutional framework. This architecture predicts affordance maps from linguistic instructions, thereby determining Regions of Interest (ROI) for grasping. Diverging from conventional two-stage grasping pipelines, works \cite{c12},\cite{c13} employ joint visual-linguistic feature learning to directly predict text-conditioned 2D grasp rectangles from RGB images. These approaches have made significant methodological contributions to language driven grasping research. However, these methods require semantically clear instructions as prerequisites and cannot handle ambiguous or complex instructions that require reasoning. In contrast, our framework is designed to integrate LLMs' commonsense and employ a multi-turn conversational interaction modality, translating language instructions of varying complexity into clear operational sequences, effectively addressing scenarios with diverse linguistic commands.

\subsection{Task-oriented Grasping}

In practical applications, robots need to dynamically select the optimal grasping strategy based on various task requirements, namely Task-Oriented Grasping (TOG). Such processes typically involve learning the affordances of objects, including the identification of potential interactive functions of various object parts, which guide the generation of grasping strategies and task execution. The affordance of objects serves as a crucial link between perception and action. Some studies \cite{c14},\cite{c15} propose learning affordances from video by observing human-object interactions, extracting the functional regions of objects, and applying them to robotic tasks. Reid et al. proposed an end-to-end affordance prediction model, Affordance-Net \cite{c16}, which first locates and classifies objects in the image, and then assigns the most relevant affordance labels to the pixels that constitute the objects. With the emergence of 3D object-part point cloud datasets, \cite{c17},\cite{c18},\cite{c19} predict the affordances of different regions in the point cloud by analyzing the contextual information and geometric features of each point in the cloud. Although these methods acquire considerable flexibility, they typically require training on large-scale datasets to gain the ability to predict object affordances. In contrast, our work implements TOG using a more streamlined approach: based on the current task context, a fine-tuned LLM determines the descriptions of target object part. The semantic cues from these descriptions are then used in conjunction with a pre-trained part segmentation model and scene depth information to segment the point clouds of the target region, effectively bypassing conventional affordance prediction steps and achieving fine-grained, part-level manipulation.

\section{METHOD}

In this section, the definition of natural language-guided robotic grasping tasks is given and describe the three key modules of LangGrasp: perception and inference, point cloud localization, and grasp pose detection.

The LangGrasp, which is proposed in this paper, is a robotic manipulation procedure in which perception, interaction, and execution are integrated. As shown in Figure \ref{fig2}. Given a scene image \textit{R}, a corresponding depth image \textit{D}, and a dialogue text \textit{H}, the aim is to generate the optimal 6-DoF grasping pose \textit{G} for the target object or object part. The perception and inference stage takes as input the image \textit{R} and the dialogue text \textit{H}, and outputs a formatted action sequence \textit{T}. The point cloud localization stage inputs the target information \textit{O} obtained from parsing the action sequence \textit{T}, and outputs a point cloud region \textit{P} containing the target object. The grasping pose prediction stage takes the point cloud region \textit{P} and action information \textit{A} as inputs and outputs the final grasping pose \textit{G}. The motion planning generates a trajectory based on \textit{G}.

\subsection{Perception and Inference Module}

In the perception and inference stage, we use the LLM (GPT-4o) for image understanding and logical reasoning, employing it as the high-level perception and interaction core of the procedure. The fine-tuned LLM, \textit{F$_{GPT}$}, is then used to generate a formatted action sequence \textit{T}. This output is derived from the current environmental information \textit{E}, as well as the clarified demand context \textit{H$_{i}$}, which is established through \textit{i} rounds of dialogue between the user \textit{C} and the LLM.
\vspace{-2pt}
\begin{equation}
    T = {F_{GPT}}\left( {E,{H_i}} \right)
    \label{EQ1}
\end{equation}
\begin{equation}
    {H_i} = {\rm{ }}\{ ({C_1},{M_1}),{\rm{ }}({C_2},{M_2}){\rm{ }}, \ldots ,{\rm{ }}({C_i},{M_i})\}
    \label{EQ2}
\end{equation}

The operation sequence \textit{T} consists of the task requirement description \textit{U} and the operational steps \textit{S$_{i}$}, with each step containing information about the action and the target object.
\begin{equation}
    {S_i} = {\rm{ }}\left( {Action,{\rm{ }}Target} \right)
    \label{EQ3}
\end{equation}
\begin{equation}
    T = \{ U{\rm{,  }}S_1{\rm{,  }}S_2{\rm{,}} \ldots ,S_{\rm{n}}\}
    \label{EQ4}
\end{equation}

\textit{T} is stored in a JSON format document. By parsing the JSON-formatted sequence \textit{T}, the target object or part at each step of the operation is extracted, along with the action information to be executed at that step. This information is then passed to the subsequent visual localization module and grasp pose detection module for the execution of the next action.

It is noteworthy that while prompt engineering can guide the generation of specific output formats through simple prompt examples, it has been revealed through our experiments that limitations still exist in the fine-grained reasoning and stability of the outputs produced by LLMs guided by prompts. Particularly when confronted with more complex language instructions, sufficiently detailed content and formatted response results are not always generated by these models. To address this, we fine-tune the LLM on a diverse set of dialogue samples with varying complexities. This enables the model to better process and execute instructions of different complexities, thereby improving its reasoning performance in grasping tasks.

\subsection{Point Cloud Localization Module}

As shown in the second part of Figure \ref{fig2}, the scene RGB image and depth image captured by the depth camera, along with the target object semantic information \textit{O$_{i}$} parsed from the perception and inference module, are received by the point cloud localization module. \textit{O$_{i}$} is used as the query target and is input into the pretrained 2D visual localization model VLPart \cite{c20}, which employs Vision Transformer and the language model BERT for multimodal information fusion, enabling simple semantic segmentation of object or part. Once the mask region of the target object is obtained, an expansion strategy is applied by sliding an appropriate dilated kernel over the generated mask region to perform dilation operations, ensuring complete coverage of the edges and surrounding background geometric information. Equation (\ref{EQ0}) defines this process. The dilation of the image is achieved by using a sliding window \textit{S} to traverse the set of pixels (\textit{x, y}) contained within the target mask image \textit{I}, resulting in the dilated image \textit{D}(\textit{I, S}).  This helps prevent unreasonable grasp poses caused by missing background information or edge noise interference. Subsequently, the expanded image is converted into a binary map, and the region of interest for grasping is identified. The depth map is transformed into a single-view point cloud using camera intrinsic parameters, and the binary map is employed for registration with the 3D point cloud, focusing on the local point cloud of the target object or part within the global point cloud. This process provides accurate positional information and geometric details of the object for the subsequent grasp pose detection.

\begin{equation}
    D(I,S)=\max\left(I(x+m,y+n)\right)\quad\forall(m,n)\in S
    \label{EQ0}
\end{equation}

\subsection{Grasp Pose Detection Module}

This stage involves the generation of appropriate 6-DoF grasping poses for the specified object of interest or part. In our previous work \cite{c21}, an end-to-end grasping pose prediction network was proposed, which takes the scene point cloud as input and outputs the 6-DoF grasping poses for the entire scene. This network is employed as the grasp module in LangGrasp. The ROI point cloud obtained during the localization stage is used as input to predict the grasp pose, which includes the rotation angle, approach distance, and confidence score. Then, the optimal grasp pose is selected based on the confidence scoring mechanism. Notably, various 6-DoF grasp pose estimation methods can be integrated into LangGrasp, providing good-expandability of our framework.

\vspace{1pt} 
\renewcommand{\arraystretch}{1.30} 
\begin{table}[ht]
    \centering
    \caption{Grasping objects dataset.}
    \label{tab1}
    \resizebox{\columnwidth}{!}{
        \begin{tabular}{c | c}
            \toprule 
            \textbf{Category} & \textbf{Items} \\
            \hline
            Tool & Hammer, Screwdriver, Spoon, Fruit knife \\
            
            Food & Banana, Apple, Carrot \\
            
            Container & Mug, Basket, Plate, Vase \\
            
            Office Supplies & Pen, Scissors, Mouse, Tape \\
            \bottomrule 
        \end{tabular}
    }
\end{table}

\begin{figure}[htbp]
    \centering
    \includegraphics[scale=0.045]{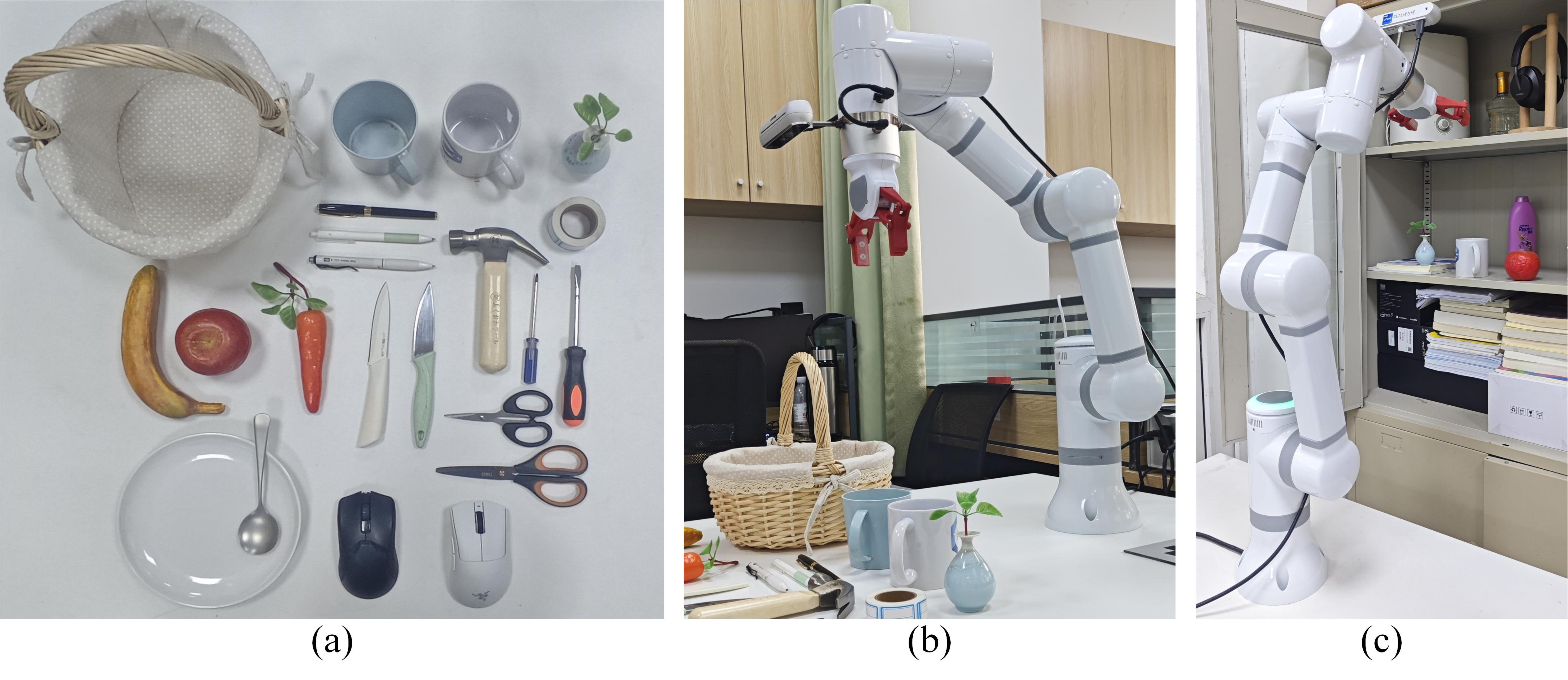}
    \caption{Experimental objects and platforms. (a) Objects dataset. (b) Desktop experimental scene. (c) Cabinet experimental scene.}
    \label{fig4}
\end{figure}

\section{EXPERIMENT}

In this section, the experimental setup is described and three experiments are conducted to evaluate our proposed method. Part A provides an overview of the dataset, experimental setup, and evaluation metrics. Part B evaluates the performance of the fine-tuned LLM in parsing language instructions, highlighting the advantages of the fine-tuning dataset. Part C examines the improvements in grasp pose quality facilitated by the point cloud localization module. Lastly, Part D demonstrates the LangGrasp framework's real-world performance.

\begin{figure}[htbp]
    \centering
    \includegraphics[scale=0.04]{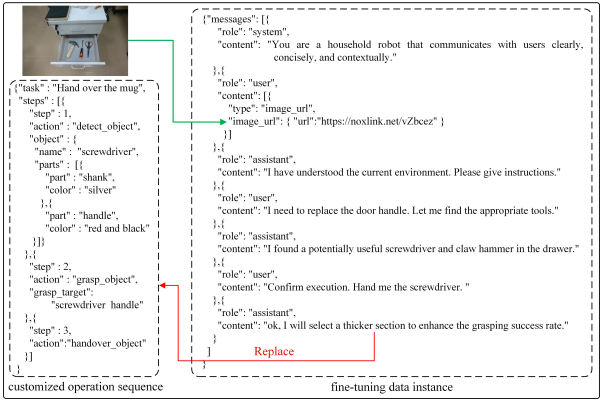}
    \caption{Collection and processing of the fine-tuning dataset.}
    \label{fig3}
\end{figure}

\subsection{Experimental Setup}

\begin{noindent}\textbf{Data Collection Setup and Fine-tuning.} (1) Image data collection: We randomly selected four categories of daily used items, including tools, food, containers, and office supplies, with three to four items per category. From a total of 22 items, 1 to 5 items were randomly placed in various scenes, such as cabinets, tables, and drawers. RGB and depth images of each scene were captured using an Intel Realsense D455 camera, with a resolution of 1280×720. (2) Dialogue data collection: Each scene image is accompanied by 1 to 3 dialogue sets with varying intentions, covering a range of language instructions. The dialogue samples were generated through interactions between users and the LLM. Prior to the dialogue, the scene image was input into GPT-4o to provide it with spatial layout and environmental information. The task information generated by GPT after a multi-round conversation is saved in a JSON file, which includes four designed basic actions: detecting objects, grasping objects, placing objects, and handing over objects. Each action records the name, features, and the order of execution steps for the target object. These standardized descriptions allow basic actions to be flexibly combined into different operation sequences, adapting to various scenarios and task requirements.\end{noindent}

\begin{table*}[ht]
    \centering
    \caption{Classification of language instructions by category.}
    \label{ta2}
    \renewcommand{\arraystretch}{1.1} 
    \begin{tabular}{m{2.1cm} m{6cm} m{6cm}}
        \toprule 
        \textbf{Instruction Type} & \hspace{1.3cm} \textbf{Description} & \hspace{2.4cm} \textbf{Examples} \\
        \hline
        \hspace{0.4cm} Simple & Single-step operation with a clear goal, no reasoning required. & 
        \begin{enumerate}
            \item Pick up the pen.
            \item Please pass me the cup.
            \item Pick up the cell phone on the table.
        \end{enumerate} \\
        \hline
        \hspace{0.3cm} Ordinary & Contains implicit information, requires contextual understanding to determine the target or task. & 
        \begin{enumerate}
            \item I am a bit thirsty.
            \item Please give me the fruit with low calories.
            \item Provide me with an appropriate tool to open the package.
        \end{enumerate} \\
        \hline
        \hspace{0.3cm} Complex & Involves multiple steps without explicitly defining specific operational targets, they are generally common that provide an overall description of the task. & 
        \begin{enumerate}
            \item Tidy up the table.
            \item Put the food into the basket.
            \item Place the tools on the table into the cabinet.
        \end{enumerate} \\
        \bottomrule 
    \end{tabular}
    \vspace{-6pt} 
\end{table*}

\begin{noindent}\textbf{Language Instruction Dataset.} Based on the complexity of natural language and reasoning requirements, we classify language instructions into three categories: simple, ordinary, and complex instructions. The classification criteria for each level are further detailed in Table \ref{ta2}.\end{noindent}

\begin{figure}[ht]
    \centering
    \includegraphics[scale=0.05]{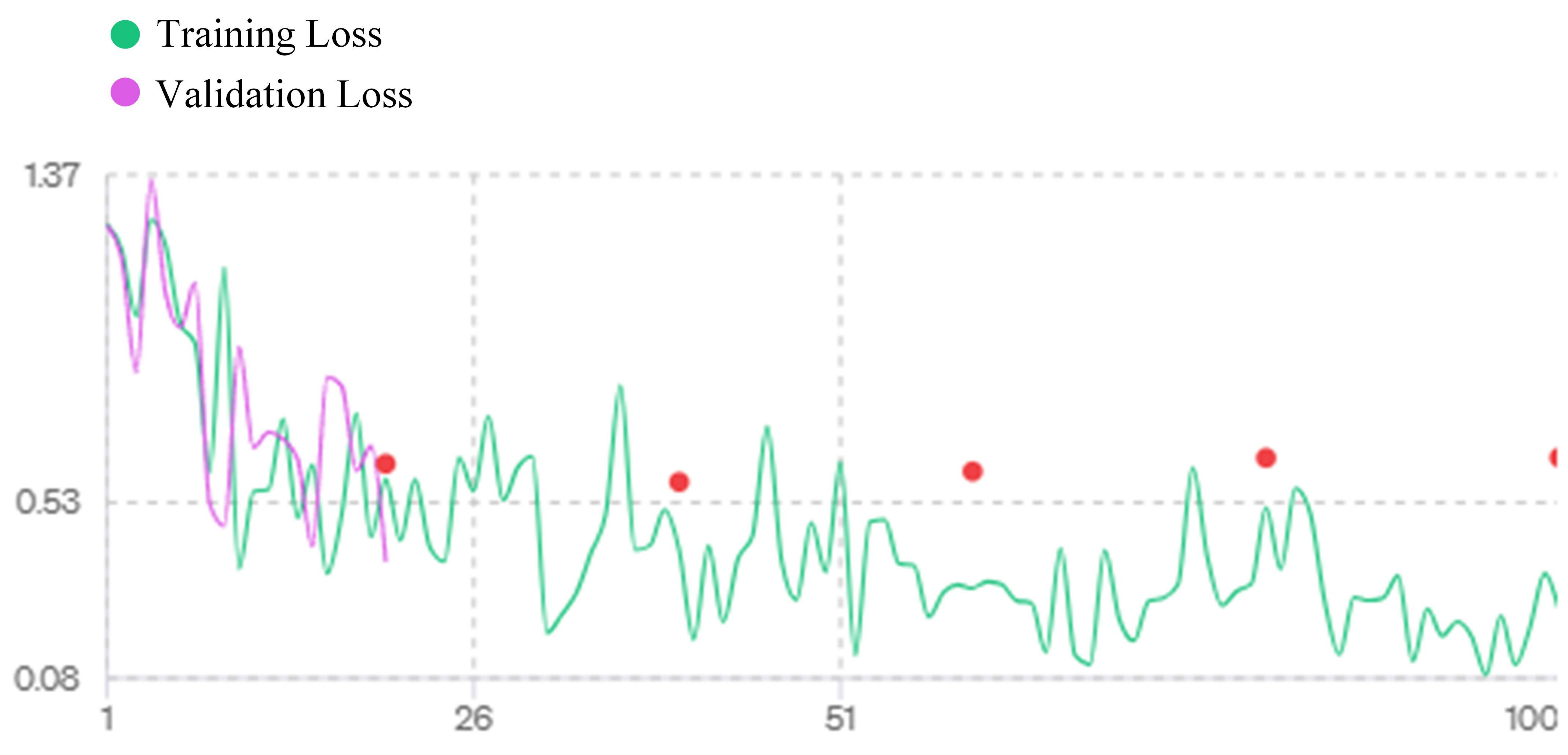}
    \caption{GPT-4o online fine-tuning process. A total of 100 training epochs were conducted, with the training loss rapidly decreasing in the early stages before stabilizing, indicating effective convergence and no significant overfitting.}
    \label{fig5}
\end{figure}

\begin{noindent}\textbf{Evaluation Metrics.} We employ the following metrics to evaluate the capabilities of our method.\end{noindent}

\textit{Semantic Understanding (SU)}. The ability to comprehend language in a manner similar to humans involves not only recognizing literal information but also inferring underlying intentions by considering context and environmental cues. For example, given the task ``I am a bit thirsty," the framework should interpret it as an indirect request and, based on environmental cues, select a water cup as the grasping target. This capability emphasizes the accuracy of semantic recognition, context processing, and understanding of environmental cues. The formula for calculating this metric is given in (\ref{EQ5}).
\begin{equation}
    SU = \frac{1}{N}\sum\limits_{i = 1}^N {{R_i}}
    \label{EQ5}
\end{equation}
here \textit{N} represents the total number of instructions, and \textit{R$_{i}$} indicates the score for the correctness of the semantics of the \textit{i-th} instruction (1 for correct, 0 for incorrect).

\textit{Structured Output (SO)}. The ability to convert unstructured text into structured data is essential. Our goal is for the output of this module to be an operation sequence in JSON format, as shown in Figure \ref{fig3}, where each field and its corresponding value are relevant to the task. The calculation formula is given in (\ref{EQ6}).
\begin{equation}
    SO = \frac{1}{N}\sum\limits_{j = 1}^N {\frac{{F_j^{{\rm{correct}}}}}{{F_j^{{\rm{total}}}}}} 
    \label{EQ6}
\end{equation}
Where $\textit{F}_{\text{j}}^{correct}$ indicates the number of correctly generated fields in the \textit{j-th} language instruction. $\textit{F}_{\text{j}}^{total}$ denotes the total number of fields generated in the \textit{j-th} language instruction (including both correct and incorrect fields).

\textit{Inference Granularity (IG)}. Fine-grained segmentation requires a hierarchical understanding of object structure for precise task execution and recognition. This metric assesses composition depth and feature identification, ensuring sufficient resolution in the output. The formula is given in (\ref{EQ7}).
\begin{equation}
    IG = \frac{1}{N}\sum\limits_{k = 1}^N {\frac{{S_k^{{\rm{correct}}}}}{{S_k^{{\rm{total}}}}}}
    \label{EQ7}
\end{equation}
Where $\textit{S}_{\text{k}}^{correct}$ denotes the correctly identified target object substructures in the \textit{k-th} language instruction. while $\textit{S}_{\text{k}}^{total}$ represents the total number of target object substructures in the same instruction.

\begin{table}[htbp]
    \centering
    \caption{Performance results of GPT-4o before and after fine-Tuning.}
    \label{tab3}
    \renewcommand{\arraystretch}{1.35} 
    \resizebox{\columnwidth}{!}{
        \begin{tabular}{c|c|c|c|c|c}
            \toprule
            {\textbf{Method}} & \textbf{\makecell[cl]{Instruction \\ \hspace{0.3cm} level}} & \textbf{SU} & \textbf{SO} & \textbf{IG} & \textbf{Overall} \\
            \hline
            \multirow{3}{*}{\makecell[cl]{Prompting \\ \hspace{0.2cm} with \\ \hspace{0.03cm} GPT-4o}} 
            & simple  & 100\% & 50\% & 40\% & 63\% \\
            & ordinary  & 90\% & 20\% & 30\% & 47\% \\
            & complex   & 85\% & 0\%  & 25\% & 37\% \\
            \hline
            \multirow{3}{*}{\makecell[cl]{Fine-tuning \\ \hspace{0.2cm} with \\ \hspace{0.03cm} GPT-4o}} 
            & simple  & 100\% & 100\% & 80\% & 93\% \\
            & ordinary  & 100\% & 100\% & 75\% & 92\% \\
            & complex   & 95\%  & 90\%  & 60\% & 80\% \\
            \bottomrule
        \end{tabular}
    }
\end{table}

\begin{noindent}\textbf{Experimental Platform Setup.} The experimental setup includes the Lebai-lm3 robotic arm kit (with a two-finger gripper), the RealSense D455 depth camera, and a laptop with a Geforce GTX 1650ti GPU. The camera is mounted on the robotic arm, and the framework is deployed on the Ubuntu 20.04. The objects used for manipulation are common daily items and tools (e.g., cups, mice, screwdrivers, hammers, fruits, etc., as shown in Figure \ref{fig4} (a)). The experimental scenarios consist of a desktop scene (Figure \ref{fig4} (b)) and a cabinet scene (Figure \ref{fig4} (c)), with 5 to 8 items randomly selected from the set and placed in each scene.\end{noindent}

\subsection{Perception and Inference Experiments}

We fine-tune the GPT-4o model using the fine-tuning dataset for grasping tasks via OpenAI's online platform. Notably, GPT-4o, trained on vast data from multiple domains, already possesses strong generalization capabilities and extensive world knowledge. This allows us to effectively standardize its output format with only a small amount of sample data during fine-tuning. Figure \ref{fig5} illustrates the process of online fine-tuning.

At the start of the task, the fine-tuned GPT-4o API is invoked with the current scene image as input, followed by iterative dialogue interactions. The task details are progressively clarified through the information obtained in each round of dialogue. When the user issues a confirmation command, such as ``Confirm execution," GPT-4o generates a final output as a JSON-encoded action sequence. This sequence is then evaluated using the pre-defined metrics.

We conducted a comparative experiment on GPT-4o before and after fine-tuning. For the version without fine-tuning, a prompt describing the current task requirements and expected response format was provided to help it understand the task background and output format. In each scene, 5 to 8 items are randomly selected, with the specific items listed in Table \ref{tab1}. Based on the selected items, 10 language instructions are designed for each level, resulting in a total of 30 instructions for each scene.

\begin{figure*}[htbp]
    \centering
    \includegraphics[scale=0.063]{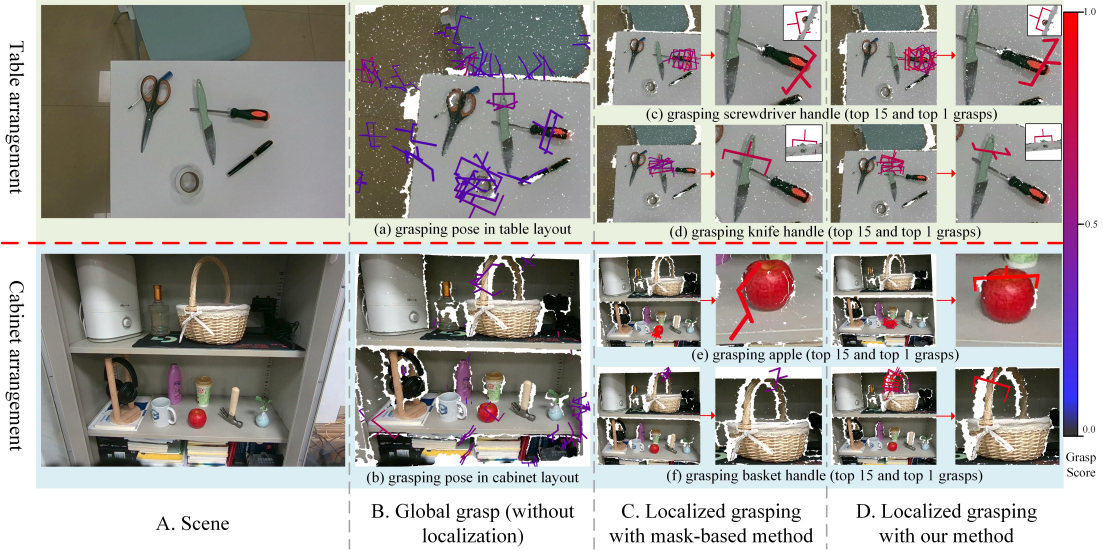}
    \caption{Visualization results of point cloud localization module experiments. Column A shows the current scene. Column B displays the global grasp poses without using point cloud localization. Column C presents the grasp pose predictions using the mask-based point cloud localization strategy, with the left and right sub-columns representing the top 15 and the top 1 grasp poses, respectively. The side-view point clouds indicate that the top 1 grasp pose results in a collision. Column D illustrates the grasp poses generated using the point cloud localization strategy proposed in this paper, with the left and right sub-columns showing the top 15 and the top 1 grasp poses, respectively. Our method not only improves the quality of the grasp poses but also eliminates the risk of collisions.}
    \label{fig6}
\end{figure*}

\begin{table*}[ht]
    \centering
    \caption{Success rates of grasping with different point cloud localization strategies. ``-'' Indicates the strategy was not applied, ``$\checkmark$'' Indicates the strategy was applied.}
    \label{tab4}
    \renewcommand{\arraystretch}{1.1} 
    \begin{tabular}{c|c|c|c|c|c}
        \toprule
        Scene & \makecell[cl]{Part/object} & \makecell[cl]{Global grasp (without \\ \hspace{0.005cm} localization strategy)} & \makecell[cl]{The mask-based \\ \hspace{0.005cm} localization \cite{c1}} & \makecell[cl]{The expansion-based\\ \hspace{0.023cm} localization (Ours)} & Success rate \\
        \hline
        \multirow{6}{*}{\makecell[cl]{Table \\ scene}} & \multirow{3}{*}{\makecell[l]{Screwdriver \\ \hspace{0.2cm} handle}} & $\checkmark$ & - & - & 0/10 \\
        
        & & - & $\checkmark$ & - & 1/10 \\
        & & - & - & $\checkmark$ & 8/10 \\
        \cline{2-6}
        & \multirow{3}{*}{\makecell[l]{Knife\\handle}} & $\checkmark$ & - & - & 0/10 \\
        & & - & $\checkmark$ & - & 0/10 \\
        & & - & - & $\checkmark$ & 6/10 \\
        \hline
        \multirow{6}{*}{\makecell[cl]{Cabinet \\ \hspace{0.01cm} scene}} & \multirow{3}{*}{\makecell[l]{Basket\\handle}} & $\checkmark$ & - & - & 0/10 \\
        & & - & $\checkmark$ & - & 3/10 \\
        & & - & - & $\checkmark$ & 7/10 \\
        \cline{2-6}
        & \multirow{3}{*}{Apple} & $\checkmark$ & - & - & 1/10 \\
        & & - & $\checkmark$ & - & 4/10 \\
        & & - & - & $\checkmark$ & 8/10 \\
        \bottomrule
    \end{tabular}
\end{table*}

The experimental results in TABLE \ref{tab3} show that fine-tuning GPT-4o outperforms prompting GPT-4o, especially in the two key metrics of \textit{SO} and \textit{IG}, where the fine-tuning method demonstrates a clear advantage. Specifically, in simple instruction scenarios, the prompting method demonstrates a high level of semantic understanding (\textit{SU} = 100\%), but its performance in structured output (\textit{SO} = 50\%) and inference granularity (\textit{IG} = 40\%) is lacking. The fine-tuning method standardizes the output format through training, enhancing the LLM's ability to capture object details, which improves \textit{SO} and \textit{IG} to 100\% and 80\%, respectively. As a result, overall performance increases from 63\% to 93\%. This effectively enhanced the model's ability to extract implicit semantics and plan multi-step operations, indicating that fine-tuning not only leverages the generalization capabilities of GPT-4o but also significantly improves its performance in specific domains with a small amount of data.

\subsection{Impact of Point Cloud Localization Strategies Experiments}

The point cloud localization strategy aims to provide key target point cloud regions for subsequent grasp pose estimation. We designed two sets of control experiments: one without local point cloud localization, referred to as global grasping, and the other using a mask-based localization strategy \cite{c1}. To validate the framework's feasibility and effectiveness, only valid mask segmentation results are considered, excluding erroneous or unavailable cases. The global grasping strategy relies on the point cloud data from the entire scene, increasing computational complexity and susceptibility to environmental clutter and interference. The mask-based point cloud localization strategy focuses on the target object or its parts, reducing interference from irrelevant information. In the grasping experiments conducted in desktop and cabinet scenes, we recorded the grasp success rates for different localization strategies. As shown in Table \ref{tab4}, without the use of a point cloud localization, the success rate for grasping the target object or its part is nearly zero. This is due to the inability to focus on the target point cloud region, leading to the prediction of grasping poses for all objects in the scene. Furthermore, when the point cloud quality is poor, erroneous grasping poses may be generated. Compared to the scenario without point cloud localization strategy, using the mask-based point cloud localization strategy improves the grasp success rates. However, in most cases, grasping still fails. The grasping network predicts poses based solely on the local point cloud within the target mask region. While this approach minimizes background noise, it also removes crucial reference information, preventing the network from determining the target object’s relative position and orientation in space. As a result, the generated grasp poses may lead to collisions. However, When using the localization strategy based on the expansion operation proposed in this paper, the grasp success rate improves significantly. The expansion operation ensures that the localized target object includes complete edge and partial background information, filtering out grasp poses that would result in collisions with the background surrounding the target. Figure \ref{fig6} presents an intuitive comparison of the 6-DoF grasping poses generated using different localization strategies.

\begin{table}[h]
    \centering
    \renewcommand{\arraystretch}{1.18} 
    \caption{Real-world interactive grasping experiments. The accuracy of the output results from each module during the execution of  different instructions is reported, along with the actual success rate of the grasping attempts. ``PaI" refers to the Perception and Inference module, ``PCL" to the Point Cloud Localization module, and ``GPD" to the Grasp Pose Detection module.}
    \label{tab5}
    \begin{tabular}{m{0.7cm} m{1cm} c m{0.5cm} m{0.5cm} c}
        \toprule
        \multicolumn{2}{c}{\diagbox{\makecell[cl]{Instruction \\ \hspace{0.03cm} level}}{Index}} & PaI & PCL & GPD & \makecell[cl]{Success \\ rate (\%)} \\
        \midrule
        \multicolumn{2}{c}{\hspace{-0.15cm}simple} & 100\% & 100\% & 93\% & 93\% \\
        \multicolumn{2}{c}{\hspace{-0.15cm}ordinary} & 100\% & 93\% & 93\% & 87\% \\
        \multicolumn{2}{c}{\hspace{-0.15cm}complex} & 87\% & 87\% & 80\% & 80\% \\
        \bottomrule 
    \end{tabular}
    \vspace{-10pt} 
\end{table}

\subsection{Real-World Interactive Grasping Experiments}

We evaluate the performance of LangGrasp through real-world interactive grasping experiments, where 5 to 8 objects are randomly selected from object dataset and placed in the experimental scene. Based on the selected items, 15 language instructions are designed for each level of complexity. The overall performance of the framework is evaluated based on the accuracy of each module's output and the actual grasp success rate. The video\footnote{\url{https://github.com/wu467/LangGrasp}} available at our github website.

The experimental results in Table \ref{tab5} show that the proposed framework performs well in real-world robotic grasping tasks, especially for simple and ordinary instructions. With the increase in instruction complexity, both the accuracy of individual modules and the actual grasping success rate exhibit certain degrees of decline. During the experiments, we observed that the primary cause of task failure was the overlap or occlusion between objects, which hindered accurate object localization and led to the failure of grasp pose generation. 
\section{CONCLUSION}

In this paper, we present LangGrasp, a novel language-guided robotic grasping framework, and construct a fine-tuning dataset for LLMs specifically designed for robotic grasping tasks. The framework leverages a fine-tuned LLM for commonsense reasoning, enabling the conversion of language instructions, regardless of complexity, into explicit action sequences. Combined with a part-level visual segmentation model, the proposed point cloud localization strategy significantly enhances the quality and accuracy of the target object’s 6-DoF grasp pose. Experimental results show that LangGrasp demonstrates robust adaptability and effectiveness in practical grasping tasks. In future work, we aim to expand the LangGrasp framework to enhance its adaptability in diverse task environments, including multi-object and similar object grasping in complex scenes, as well as real-time grasp adjustment in dynamic settings.

\addtolength{\textheight}{-12cm}   





\bibliographystyle{IEEEtran}  

\bibliography{root}

\end{document}